\pdfoutput=1

\RequirePackage{snapshot}
\documentclass[11pt]{article}
\usepackage{times,latexsym}
\usepackage{url}
\usepackage{listings}
\usepackage{multirow}
\usepackage{caption}
\usepackage{subcaption}
\usepackage[T1]{fontenc}
\usepackage{color,soul}
\usepackage{hyperref}
\usepackage{amsmath,amsfonts,bm}
\usepackage{dialogue}
\usepackage[export]{adjustbox}
\usepackage[capitalize]{cleveref}
\crefname{section}{$\mathsection$}{$\mathsection\mathsection$}
\Crefname{section}{$\mathsection$}{$\mathsection\mathsection$}
\usepackage{xspace,mfirstuc,tabulary}
\usepackage{acl}

\usepackage{booktabs}
\usepackage{graphicx}
\newcommand{\shortparagraph}[1]{\paragraph{#1}}

\title{Identifying Self-Disclosures of Use, Misuse and Addiction in Community-based Social Media Posts}

\author{
Chenghao Yang$^{1,*,\dagger}$, Tuhin Chakrabarty$^{2,*}$, Karli R Hochstatter$^{3,\dagger}$, Melissa N Slavin$^{4,\dagger}$, 
\\ 
\textbf{Nabila El-Bassel}$^{5}$, \textbf{Smaranda Muresan}$^{2,6}$
\\ $^1$University of Chicago $^2$Department of Computer Science, Columbia University
\\ $^3$Friends Research Institute $^4$Fairleigh Dickinson University
\\ $^5$School of Social Work, Columbia University
$^6$Data Science Institute, Columbia University
\\ \texttt{chenghao@uchicago.edu, \{tc2896, sm761\}@columbia.edu}
}

\date{}

\begin{document}
\maketitle
\def\thefootnote{$*$}\footnotetext{Equal Contribution.}\def\thefootnote{\arabic{footnote}}
\def\thefootnote{$\dagger$}\footnotetext{Work done when authors were at Columbia University. }\def\thefootnote{\arabic{footnote}}

\begin{abstract}

In the last decade, the United States has lost more than 500,000 people from an overdose involving prescription and illicit opioids\footnote{\url{https://www.cdc.gov/drugoverdose/epidemic/index.html}} making it a national public health emergency~\citep{us2017hhs}.
Medical practitioners require robust and timely 
tools that can effectively identify at-risk patients. 
Community-based social media platforms such as Reddit allow
self-disclosure for users to discuss otherwise sensitive drug-related behaviors.
We
present 
a moderate size corpus of 2500 opioid-related posts from various subreddits labeled with six different phases of opioid use: \textit{Medical Use, Misuse, Addiction, Recovery, Relapse, Not Using}. For every post, we
annotate span-level extractive explanations and crucially study their role both in annotation quality and model development.\footnote{The codebase and dataset specification are available at \url{https://github.com/yangalan123/OpioidID}.}
 We evaluate several state-of-the-art models in a supervised, few-shot, or zero-shot setting. Experimental results and error analysis show that identifying the phases of opioid use disorder is highly contextual and challenging.
However, we find that using explanations during modeling leads to a significant boost in classification accuracy
demonstrating their beneficial role in a high-stakes domain such as studying the opioid use disorder continuum. 

\end{abstract}

\section{Introduction}
Extensive ongoing overuse of opioid medications, both from medical prescriptions and from illegal sources has led to a major
public health crisis~\citep{degenhardt2019global, krausz2021opioid}.
There have been a total of 103,664 drug overdose deaths in the US in the 12-month period ending April 2022.\footnote{\url{https://www.cdc.gov/nchs/nvss/vsrr/drug-overdose-data.htm}}
For individuals with opioid use disorder (OUD), targeted interventions need to be developed to better capture individuals' transitions at critical junctures (e.g., use to misuse; misuse to addiction; recovery to relapse)~\citep{park2020situating}.

\begin{table}[]
\small
\resizebox{\columnwidth}{!}{
\renewcommand{\arraystretch}{1.0}
\centering
\begin{tabular}{|p{0.86\columnwidth}|}
\hline
I'm 18m and I've been taking norcos since I was 16 but just on and off. \textcolor{red}{\textbf{\hl{Starting this year I've been taking it every day basically and now I'm tired of it. I still get high
so ig my addiction isn't that bad}}} as others but I don't want to get to that point. I'm tired of chasing the high. I've spent at least 3k on norcos this year and \textcolor{red}{\textbf{\hl{I can't control }}} \textcolor{red}{\textbf{\hl{myself}}}. I try to go a day sober but my mind is telling me I need 
and then withdrawals starts 
[...]
\\ \hline
\end{tabular}
}
\caption{\label{excerpt}A self-disclosure from a user on Reddit going through the cycle of Opioid Addiction.} 
\end{table}

Due to their anonymous and real-time participation, community-based social media platforms such as Reddit, have been used by researchers to understand issues around mental health self-disclosure \cite{Choudhury2014MentalHD}, suicide among youth \cite{Sumner2019TemporalAG},
marijuana regulations \cite{park2017tracking}, drug community analysis~\cite{bouzoubaa2023exploring} and Covid-19 impact on people who use opioids \cite{el2022harnessing}.
We choose Reddit for our research,
specifically the popular opioid-related subreddits \textit{r/Opiates, r/OpiatesRecovery} as well as \textit{r/drugs} to collect our data (\cref{collection}).
Our research focuses on
predicting the presence of self-disclosures related to OUD phases in users' Reddit posts (refer to \cref{excerpt} for an example). 
This task is critical in providing healthcare professionals and social workers with automated tools for detecting OUD indications in social media posts. Accurate identification of such self-disclosures can enable more effective, targeted interventions for individuals suffering from OUD, as supported by prior research \cite{acion2017use, park2020situating, hasan2021machine}.
Our goal is to establish an annotation framework based on addiction and substance use research, categorizing behaviors like Medical Use, Misuse, Addiction, Recovery, and Relapse. We also seek to demonstrate the effectiveness of recent NLP advancements, especially through the application of explanations and text-to-text models, in accurately identifying self-disclosures within the OUD continuum. We offer three primary contributions:
\begin{itemize}
\item 
{\it An
annotation scheme amenable for both expert and novice annotations of self-disclosures.}
The proposed scheme has three characteristics: 1) is grounded in research on addiction and substance use
2) aims to focus on self-disclosure of OUD phases by including a category Not Using that applies to posts that are not discussing the author's OUD experience; and 3) aims to provide reliable annotations by both experts and novices (\cref{sec: data}). 
 \item 

{\it High-quality dataset annotated with class labels and text explanations
using expert and novice annotators.} Human annotations are essential, both to ensure that the NLP models can accurately learn to identify the various OUD phrases, and as an upper bound on the expected model performance. Towards this, we employ both substance use research experts and skilled crowd-workers to annotate our data based on our scheme (\cref{collection}). 
To ground annotators' decisions towards a particular label, we also asked them to highlight the minimum span from the input that acts as an explanation for their chosen category/label. 
 \item 

{\it Thorough experimental setup of zero-shot, few-shot, and supervised models with insights into the role of explanations for model performance, the impact of label uncertainty, and intriguing properties of users' self-disclosure. 
}
Our experiments demonstrate that: 1) the model performance improves significantly 
when trained/prompted with explanations. A further ablation study on 
human-annotated explanations versus machine-generated 
explanations confirms that the quality of explanations is key to 
such improvement;
2) smaller models
fine-tuned on our novice-annotated data with explanations works best,  
surpassing zero-shot
and few-shot large models, including GPT-4,  
by a large margin (\cref{sec: exp}); 3) an ablation study taking into account label uncertainty sheds light on model errors for cases where humans agree or disagree on the label; 4) our error analysis shows preliminary insights in understanding users' self-disclosure (\cref{sec: error_analysis}).

 \end{itemize}

\section{Data}
\label{sec: data}
\subsection{Data Collection and Annotation}
\label{collection}

\shortparagraph{Data Source} One of the greatest challenges in building models that are capable of identifying the appropriate category for opioid usage is the lack of publicly available large-scale datasets. Social media platforms such as Reddit often provide social support
for people who use opioids, while allowing for anonymity when discussing stigmatized behaviors \cite{pandrekar2018social, Bunting2021SociallysupportiveNA}.
We collect data from the popular opioid subreddits, \textit{r/Opiates} and \textit{r/OpiatesRecovery} as well the \textit{r/drugs} subreddit.
Since \textit{r/drugs} can contain posts related to other drugs, we only select posts that are labeled with a flair(tag) ``opioids" by the moderator.

\begin{table*}[t]
\small
\centering
\renewcommand{\arraystretch}{1.0}
\begin{tabular}{@{}cl@{}}
\toprule
Medical Use
& \begin{tabular}[c]{@{}l@{}}\underline{\textbf{Oxycodone for wisdom teeth removal }}\textcolor{red}{\textbf{\hl{I just got 4 wisdom teeth plus another tooth in}}}\\ \textcolor{red}{\textbf{\hl{my palette removed and got prescribed 1 or 2 5mg tablets of oxy (Endone) each time}}}. \\He recommended to avoid it if I could since I'm 43kg and have no tolerance. [...]
\end{tabular}   \\ \midrule
Misuse & \begin{tabular}[c]{@{}l@{}}\underline{\textbf{Oxy nod but no euphoria? }} Hi everyone, \textcolor{red}{\textbf{\hl{I tried oxy for the first time a few weeks back}}}\\ \textcolor{red}{\textbf{\hl{snorting a prolonged 20mg tablet and felt pretty good}}}. Wednesday I dropped 9 of the \\5mg capsules over a couple hours and was nodding strongly [...]
\end{tabular} \\ \midrule
Addiction       &   \begin{tabular}[c]{@{}l@{}}\underline{\textbf{Well y'all were right. }} The sickness came. And is the worst i've ever experienced. Took \\subs, went into pwd  accidentally and jump started the methadone sickness. \textcolor{red}{\textbf{\hl{I am to the}}} \\\textcolor{red}{\textbf{\hl{point that I just have to get off this godforsaken mountain and go back to my ex and }}}\\\textcolor{red}{\textbf{\hl{get back in the clinic bc at this rate i'm afraid i'm gonna end up killing myself}}}.[...].
\end{tabular} \\ \midrule
Recovery       & \begin{tabular}[c]{@{}l@{}}\underline{\textbf{It's my birthday!} \textcolor{red}{\textbf{\hl{One year off opiates}}}}  It's been 365 days since I decided to take back\\ control of my body. I was highly dependent and addicted to prescribed opiates [...]
\end{tabular} \\ \midrule
Relapse      &  \begin{tabular}[c]{@{}l@{}}\underline{\textbf{So high. 18 hours later. Still so high. }}So I'm pissed at myself. \textcolor{red}{\textbf{\hl{I was clean from heroin }}}\\\textcolor{red}{\textbf{\hl{for 11 months and last night I did some}}}. And for no reason too [...] 
\end{tabular}                                                       \\ \midrule
Not Using 
& \begin{tabular}[c]{@{}l@{}}\underline{\textbf{Partners of an Opiate addict in recovery }}How do you guys do this? I feel like I am \\having an incredibly hard time "moving on". \textcolor{red}{\textbf{\hl{I have nightmares of my partner oding,}}} \\\textcolor{red}{\textbf{\hl{dying, and pretty much anything else that involves drug use}}}. I over analyze everything [...]
\end{tabular}                                                                                                                                                                                                                                                                                                                                                                                                                                                                                                                \\ \bottomrule
\end{tabular}
\caption{\label{data} Example for each
Opioid Usage category. The underlined bold text represents the title of each post. Highlighted text
represents salient spans annotated by humans as explanations for the label.}
\end{table*}

\shortparagraph{Anonymization and Data Preprocessing} 
To remove any personal identifying information (PII) that users might divulge in their posts (e.g., emails) and broken characters, we use \textit{cleantext}\footnote{\url{www.github.com/prasanthg3/cleantext}} to preprocess raw social media posts. In addition, we manually investigated
all samples prepared for annotation to make sure PII will not be exposed to annotators, and thus will not be released in the final dataset.  After that, 
we check whether each post is of reasonable length (title + text), and
filter the preprocessed posts having a length of less than $10$ words (%
or more than $200$ words for easier annotation).
We sample $600$ posts for expert annotation and $2,250$ posts for novice 
annotation.\footnote{Domain experts are postdoctoral and advanced doctoral students working in substance abuse research. We use MTurk for novice annotation. }

\shortparagraph{Annotation Guidelines} 
To ensure the annotation quality, we worked closely with substance use research experts to develop comprehensive and precise annotation guidelines for different phases of opioid use. OUD has been recognized as a chronic, relapsing disorder in which individuals may begin at one stage, remain in that stage, gradually or rapidly advance to another stage, enter recovery, return to use, or even skip stages~\citep{volkow2007science}. 
For this study, we adopted frequently used classifications to assign each post a stage in the continuum: Medical Use, Misuse, Addiction, Recovery, and Relapse \citep{nida2007drugs, smith2013classification, hanson2013tweaking, hanson2013exploration, chan2015canary, anderson2017using, phan2017enabling, hu2019ensemble}. Our definitions for Medical Use, Misuse, and Addiction come from the systematic review~\citep{smith2013classification}, and our definitions for Recovery and Relapse come from National Institute on Drug Abuse guideline~\citep{nida2007drugs}.
We also built a list of keywords, representative samples and FAQs to clarify the project background, ethical considerations, and how to handle uncertain cases. The guidelines aim to understand the opioid use experiences of \emph{the author of the post} (self-disclosures). Thus,
we introduced also a category of 'Not Using' that includes discussion about someone else who uses opioids or general questions about opioids, without evidence of use. \cref{app: annotation_guideline} shows the definitions for each category and some examples of expert-authored FAQs for clarification. \cref{tab:stat} shows the distribution of OUD categories in the annotation data.

\begin{table}[!ht]
\small
\centering
\renewcommand{\arraystretch}{0.9}
\begin{tabular}{ccc}
\toprule
Category      & Novice & Expert  \\ \midrule
Misuse      & 22.10 & 20.0  \\ 
Addiction   & 29.15 & 12.53 \\ 
Recovery    & 18.89 & 25.49 \\ 
Relapse     & 4.65  & 3.96  \\ 
Medical Use & 7.05  & 3.52  \\ 
Not Using   & 18.17 & 34.51 \\ 
\bottomrule
\end{tabular}
\caption{Distribution (\%) of OUD categories 
in novice- and expert-annotated data.}
\label{tab:stat}
\end{table}

\shortparagraph{Expert Annotation}
To build the expert evaluation dataset, we invited 4 substance use research experts to annotate $600$ posts and paid them at a rate of \$20/hour. To accommodate the experts' available timeslots, we split the posts into two equal batches and asked the experts to annotate the text and title of the post with both the label and the explanation. All four experts annotated the first batch. For label annotation, the inter-annotation agreement (IAA)
was $0.46$ Fleiss' kappa~\citep{fleiss1971measuring}, indicating ``moderate agreement''. Only two experts were available to annotate the second batch, and the IAA was $0.62$ Cohen's kappa~\citep{cohen1960coefficient}, indicating ``substantial agreement''.
We filtered the posts that did not obtain majority agreement, obtaining an
expert-annotated dataset of $455$ posts. We further split the dataset into $13$ samples
as in-context samples of few-shot prompting (\cref{sec:exp-few-shot}) and $442$ samples for testing.

\label{crowd}
\shortparagraph{Novice Annotation} Expert annotation, while being more accurate and trustworthy, is not feasible for scaling the process beyond a few hundred posts. Hence, we aimed to leverage
novice annotators using Amazon MTurk. However, to obtain reliably annotated data, we need to ensure that novices are qualified and trained.
Thus, we first conducted a qualification test where we recruited a total of 85 crowd-workers from the USA with a 98\% success rate and asked them to annotate 250 randomly selected instances from the expert-labeled set. We qualified only 10 crowd-workers who obtained >60\% accuracy in the qualification phase. In addition, for cases of disagreement with the experts, we further trained the novice annotators by providing them with follow-up explanations. We paid them \$15/hour, which is in accordance with the minimum wage in the USA. Every post is labeled by three qualified novice annotators. We labeled 2,250 posts and kept 2,086 for which we could obtain a majority vote label. 
We split this set into 1,936 for training and 150 for testing. IAA was 0.47  based on Fleiss' kappa~\cite{fleiss1971measuring} (``moderate agreement'').

\label{explanation}
\shortparagraph{Explanation Annotation} Along with providing a  label we also asked annotators to identify the minimum salient span from the text that justifies their decision towards labeling a post to a certain category.
For cases where we have a majority vote and use the corresponding label as gold, we have to decide what explanation to include. We computed the max overlapping substring between the annotators' explanations.
When the max overlapping substring is very short (typically $<10$ characters), we chose the longest explanation
whose annotated label matches the majority vote label.
For 63\% of cases, there is significant overlap among annotators' selected explanation spans, while for 37\% of cases the longest explanation is selected. 
\cref{data} shows post examples in each category/label along with their annotated span-level explanations.

\subsection{Disagreements in Annotation}
\label{sec: disagreement}

\begin{table}[]
\small
\resizebox{\columnwidth}{!}{
\renewcommand{\arraystretch}{1.0}
\centering
\begin{tabular}{|p{0.86\columnwidth}|}
\hline
\begin{tabular}[c]{@{}p{0.86\columnwidth}@{}}\textcolor{red}{\textbf{\hl{Is it safe to mix 20mg oxy with a few standard drinks?}}} Seen online it's dangerous but I don't trust a lot of those harm reduction websites, most of them are whack.
EDIT: \textcolor{blue}{\textbf{\hl{I didn't do it don't worry, thanks for the info}}}\end{tabular} \\ \hline
\end{tabular}
}
\caption{An annotation example demonstrating the role of explanation annotations for understanding annotator disagreement: the red is associated with ``Misuse'' and the blue with ``Not Using''.  }
\label{tab: different_label_different_rationale}
\end{table}
\shortparagraph{Expert-Novice Disagreement} During the qualification test, we observe a consistent labeling disagreement between our qualified novice annotators
and the expert annotators (The confusion matrix is shown in \cref{app:heatmap}).
The main disagreement between experts and novices are 
between \textbf{``Addiction'' - ``Recovery''} (22.35\%), \textbf{``Not Using'' - ``Misuse''} (19.35\%), \textbf{``Addiction'' - ``Misuse''} (12.90\%),  \textbf{``Medical Use'' - ``Misuse''} (10.75\%) and \textbf{``Recovery'' - ``Relapse''} (8.60\%). 
\shortparagraph{Novice-Novice Disagreement.} Even though we reach a majority vote
for 2086 posts,
an individual worker can still disagree on the collective label. Looking at these disagreements can help us better understand the difficulty of this task and the uncertainty in the annotated dataset. In total, $1165$ out of $2086$
(56\%) posts in our final novice-annotated datasets fall in  this category. The top-5 disagreements happen between \textbf{``Addiction''-``Misuse''} (34.84\%), 
\textbf{``Recovery''-``Addiction''} (15.71\%),
\textbf{``Not Using''-``Addiction''} (12.27\%), 
\textbf{``Not Using''-``Misuse''} (9.78\%) and \textbf{``Not Using''-``Recovery''} (7.38\%). 
This inherent uncertainty may inject wrong inductive bias into models, which we discuss in \cref{sec: error_analysis}.

A closer look at some examples of disagreement in annotations shows that selected explanations could shed some light. For example, \cref{tab: different_label_different_rationale}, shows an example of disagreement between Misuse and Not Using, where the annotators selected two different explanations for the labels. 

\section{Modeling Strategies}
\begin{table}[]
\small
\centering
\renewcommand{\arraystretch}{1.0}
\begin{tabular}{|p{0.94\columnwidth}|}
\hline
\begin{tabular}[c]{@{}p{0.9\columnwidth}@{}}Given the following title, text, and explanation from the text, please identify the appropriate opioid usage category among the following types: {\color{blue}'Medical Use', 'Misuse' ,'Recovery', 'Relapse', 'Addiction', 'Not Using'.}\\\\ 
Title: \{\{title\}\}\\  
Text:  \{\{text\}\}\\ 
Explanation:  \{\{explanation\}\}
\end{tabular} \\ \hline
\end{tabular}
\caption{Zero-shot instructional prompts for T0pp for
\textit{w/ Explanation} setting}
\label{zeroshot}
\end{table}

As OUD status prediction is a high-stakes task with limited labeled data, we consider three different settings, gradually increasing the number of labeled data required to mimic real-world application scenarios: zero-shot, few-shot, and supervised learning. 
To understand the effectiveness of annotated span-level explanations, we conduct two experiments for each setting: i) \textit{w/o Explanation}: where the explanation is not included in the input, and ii) \textit{w/ Explanation}: otherwise.

\shortparagraph{Zero-Shot} 
We first consider the extreme application scenario when zero training data is given.
In order to measure zero-shot performance on our dataset, we
prompt the widely-used instruction-tuned T0pp \cite{sanh2022multitask} model for our task. 
The prompt with instructions are demonstrated in \cref{zeroshot}.%
\footnote{We only presented \textit{w/ Explanation} case to save space.}
We use greedy search to generate the labels, then use exact match to compute accuracy after lowercasing both the output and label.

\begin{table}[]
\small
\centering
\begin{tabular}{|p{0.85\columnwidth}|}
\hline

\begin{tabular}[c]{@{}p{0.85\columnwidth}@{}}Given the following title and text, please identify the appropriate opioid usage category among the following types:  {\color{blue}'Medical Use', 'Misuse', 'Recovery', 'Relapse', 'Addiction', 'Not Using'.} 
{\color{orange} Please provide an explanation for your answer by extracting the relevant span from the text that justifies your choice.}
\\\\
\{\textcolor{red}{13 in-context samples with the format below}\} \\

Title: \{\{title\}\}\\ 
Text:  \{\{text\}\}\\ 
Label:  \{\{label\}\}\\ 
Explanation:  \{\{explanation\}\}\end{tabular} \\ \hline
\end{tabular}

\caption{Few-shot instructional prompt for GPT-3 for \textit{w/ Explanation} setting.}
\label{tab: gpt-3-prompt}
\vspace{-8pt}
\end{table}

\shortparagraph{Few-Shot}
\label{sec:exp-few-shot}
Now we relax the dataset size limitation to allow the few-shot setting. 
 We use the GPT3-Davinci-002 model~\citep{brown2020language} and GPT-4~\cite{OpenAI2023GPT4TR} for the few-shot learning method. Our prompts
begin with the task instruction followed by 13 expert-annotated samples for in-context learning.\footnote{See \cref{app:in-context-samples} for these in-context samples and the detailed explanations for selecting these examples. }
For in-context learning \textit{w/ Explanation}, we place the explanation on a line after the answer, preceded by ``Explanation:"\cite{lampinen2022can}. \cref{tab: gpt-3-prompt} shows an example prompt.\footnote{We only show \textit{w/ Explanation} case to save space.} In this way, the evaluation can be performed regardless of whether explanations are provided in the prompt or not.\footnote{Necessary post-processing during evaluation for GPT-3/4 output normalization is detailed in \cref{app:post_processing_gpt3}.}%

\begin{table*}[]
\centering
\resizebox{0.7\textwidth}{!}{%
\renewcommand{\arraystretch}{1.0}
\begin{tabular}{@{}cccccccc@{}}
\toprule
\multirow{2}{*}{Mode}            & \multirow{2}{*}{Test Set} & Zero-Shot & \multicolumn{2}{c}{Few-Shot} & \multicolumn{3}{c}{Supervised}                                    \\ \cline{3-8} 
                                 &                             & T0pp      & \multicolumn{1}{c}{GPT3} & \multicolumn{1}{c}{GPT4}    & \multicolumn{1}{c}{DeBERTa} & \multicolumn{1}{c}{T5-3B} & T5-11B \\ \hline
\multirow{2}{*}{w/o Explanation} & Expert                      & {48.9 / 46.9}      & 62.2 / 57.1 & \multicolumn{1}{c}{55.4 / 50.2}    & \multicolumn{1}{c}{67.6 / 65.7}    & \multicolumn{1}{c}{63.5 / 61.2}  &  \textbf{71.4} / \textbf{70.4}      \\ 
                                 & Novice                 & 62.0 / 60.8      & 66.4 / 65.4 &  \multicolumn{1}{c}{63.3 / 60.0}    & \multicolumn{1}{c}{74.0 / 74.4}    & \multicolumn{1}{c}{72.7 / 71.4}  & \textbf{80.9} / \textbf{81.5}       \\ \hline
\multirow{2}{*}{w/ Explanation}   & Expert                      & 48.9 / 47.4     & 61.1 / 54.5 & \multicolumn{1}{c}{63.2 / 59.1}    & \multicolumn{1}{c}{{73.8 / 72.6}}    & \multicolumn{1}{c}{{64.4 / 65.5}}  & \textbf{76.6} / \textbf{77.0}        \\ 
                                 & Novice                 & {62.7 / 58.5}      & {66.9 / 65.9} & \multicolumn{1}{c}{67.3 / 64.8}     & \multicolumn{1}{c}{{81.3 / 81.9}}    & \multicolumn{1}{c}{{78.7 / 77.3}}  & \textbf{84.0} / \textbf{84.0}       \\ \hline
\end{tabular}
}
\caption{Performance of different models on expert and novice-annotated test data in a zero-shot/few-shot/supervised setting. \textit{w/o Explanation} and \textit{w/ Explanation} models refers to the setting where \textit{Explanations} are excluded or included as part of the input. Results are presented in "Accuracy/F1" format. }
\label{tab:main_exp}
\end{table*}

\shortparagraph{Fully Supervised}
All of our training data comes from the novice-annotated set, while our test sets consist of expert or novice-annotated data.  Our training data consists of 1936 examples, while our test sets consist of 442 expert-annotated examples and 150 novice-annotated examples. We consider two modeling variants: Masked Language Models (MLM) (as it is often used in traditional fine-tuning) and Generative Language Models (GLM) (as it is often used in instruction-tuning).

For MLM, we fine-tune DeBERTa-v3-large~\citep{he2021debertav3} on our training data.%
For input formatting, we use ``[title] {TITLE} [text] {TEXT} [Rationale] {RATIONALE}'' as the input for \textit{w/ Explanation} settings and use ``[title] {TITLE} [text] {TEXT}'' to train models under \textit{w/o Explanation} settings. The token in square brackets (e.g., ``[title]'') are special tokens and the tokens in all-caps
(e.g., ``{TEXT}'') are actual text fields for each post. 
For GLM,
 we fine-tune T5-3B and T5-11B models ~\cite{raffel2020exploring}.  
We use the same instruction as input to the encoder for a given title, text and optionally explanation as the ones we used for zero-shot setting (see \cref{zeroshot}).
The decoder generates the textual label autoregressively. More implementation details for fine-tuning can be found in \cref{sec:fully_supervised_fine-tuning_detail}. 

\section{Experiments}
\label{sec: exp}
\cref{tab:main_exp} summarizes our experimental results under three different learning settings (zero-shot, few-shot, and fully supervised) across two different modes i) \textit{w/o Explanation} when only the Title and Text are a part of our input during training and testing, and ii) \textit{w/ Explanation} when along with Title and Text, gold human annotated explanations are a part of our input during training and testing. We show results on both the expert-annotated test set and the novice-annotated test set. As the OUD category distribution in our dataset is unbalanced, we report both accuracy and macro F1 scores in \cref{tab:main_exp}. We find for the model-wise comparison, there is little difference in using accuracy or F1.

We highlight several takeaways. First, adding explanations helps the models both on expert and novice-annotated data (except for T0pp and GPT3 on Expert data), particularly in few-shot and fully supervised settings. In \cref{explanations}, we will show additional experiments to study the role of explanations and their quality for model predictions. Second, supervised learning with small models
outperform
few-shot methods with larger models including GPT-4 by a large margin, on both expert and novice evaluation datasets, even if the training data is novice-annotated. T5-11B is the best overall model. While our training data is not annotated by experts, the quality of the data is still high. The accuracy on the expert evaluation set for a random baseline would be 17\%, while a majority baseline would be 35\%, which is significantly lower than 71.4\% or 76.6\% for the T5-11B model performance \textit{w/o Explanation} and \textit{w/ Explanation}, respectively. Moreover, we notice that the performance gap between expert-annotated test data and novice-annotated test data is reduced using supervised models. A closer look at GPT-4 errors shows that GPT-4 is particularly struggling with the ``Not Using'' category, which covers a diverse range of topics that can look very different from posts in other categories, and more analysis on this category will be further studied in \cref{sec: error_analysis}. Third, model capabilities improve with scale under the same family in a supervised setting. The T5-11B model, on average, is about $8.4$ points better than the T5-3B model in accuracy and $9.3$ points better in F1.
However, when models belong to different families (i.e., Generative vs. MLM), the scaling law might not hold as the DeBERTa-v3-large model (1.5B) outperforms T5-3B across both settings (\textit{w/o and w/ Explanation}).

\section{The Role of Explanations} \label{explanations}

\begin{table}[]
\centering
\resizebox{0.8\columnwidth}{!}{%
\renewcommand{\arraystretch}{1.0}
\begin{tabular}{@{}cccc@{}}
\toprule
     Explanation             &  Test Set      & T5-11B          & DeBERTa          \\ \midrule
\multirow{2}{*}{Gold}   
                  & Expert           & \textbf{76.6}   & 73.8             \\
                  & Novice           & \textbf{84.0}   & 81.3             \\ \midrule
\multirow{2}{*}{Silver} 
                  & Expert           & 70.3            & 70.3             \\
                  & Novice           & 78.0            & 78.0             \\ \midrule
\multirow{2}{*}{Random}  
                  & Expert           & 69.6 ($\pm$ 1.3) & 65.8 ($\pm$ 1.1)  \\
                  & Novice           & 68.3 ($\pm$ 1.7) & 67.9 ($\pm$ 1.4)  \\
\bottomrule
\end{tabular}
}
\caption{Accuracy of T5-11B and DeBERTa \textit{w/ Explanation} model on expert and novice annotated test sets by varying the quality of explanations. We can observe the importance of including gold explanations.} 
\label{tab:explquality}
\end{table}

To test the quality and helpfulness of the annotated explanations on model prediction, we conduct three different experiments using our two best-performing models trained \textit{w/ Explanation} (T5-11B and DeBERTa).
All these experiments are conducted at inference time on top of a model fine-tuned on <title, text, $E^{\text{gold}}$>. For convenience,  from here on, we will refer to this model as M1. 

\shortparagraph{Gold Explanations at Inference.} In the first experiment, we use the gold explanations from our test sets (expert and novice).
In particular, during inference, we prompt the two best-performing models (T5-11B and DeBERTa) with an input that consists of <title, text, $E^{\text{gold}}$>.  \cref{tab:explquality} shows that models that use gold explanations at inference time are the best. We analyze whether the explanation contains words that refer to the label (e.g., addiction or addicted), a problem referred to as \textit{leakage} \cite{sun2022investigating}.
We notice that there is 5.6\% leakage on expert-annotated test data and 8\% leakage on novice-annotated test data, which means that most of our annotated explanations do not give away the label easily.

\shortparagraph{Silver Explanations at Inference.} In a real-world setting, it is not possible to expect gold explanations at inference time. Thus, in this setting, we investigate whether model-generated explanations can still be helpful for final label prediction. 
Prior works in explainability have trained two types of models: 1) \textit{Pipeline} model, which maps an input to an explanation (I $\rightarrow$ E), and then an explanation to an output (E $\rightarrow$ O); and 2) \textit{Joint Self Explaining} models that map an input to an output and explanation (I $\rightarrow$ OE). The latter has been shown to be more reliable \cite{wiegreffe2021measuring}.
Thus, we first train a T5-11B model (M2) that can jointly generate <{label}, {explanation}> given any <{title}, {text}>. At inference time, we first generate a silver explanation ${E}_{i}^{\text{silver}}$ by prompting M2 with a given <{title}$_{i}$, {text}$_{i}$> from the test set. We then prompt M1 with <{title}$_{i}$, {text}$_{i}$, ${E}_{i}^{\text{silver}}$> to generate label$_{i}$. While these explanations are not as high quality as gold explanations, they still outperform random explanations. It should be noted that the goal of this paper is not to build models that facilitate extracting accurate explanations. However, such models
might improve the silver quality explanations and thereby improve overall classification results. We leave this for future work. 

\shortparagraph{Random Explanations at Inference.} As a baseline,
we use a randomly selected sentence from the post as the explanation. We repeat the random selection for five random seeds and report the mean and standard deviation of these five runs (\cref{tab:explquality}).
Both silver and gold explanations
outperform the random explanation baseline,
indicating the need
for
informative, high-quality explanations.

\begin{table*}[]
\centering
\resizebox{0.65\textwidth}{!}{%
\begin{tabular}{@{}ccccccc@{}}
\toprule
\multirow{2}{*}{Setting} & \multicolumn{2}{c}{Not Using - Misuse} & \multicolumn{2}{c}{Recovery - Addiction} & \multicolumn{2}{c}{Not Using - Addiction} \\ \cmidrule(l){2-7} 
 & $\rightarrow$ & $\leftarrow$ & $\rightarrow$ & $\leftarrow$ & $\rightarrow$ & $\leftarrow$ \\ \cmidrule(r){1-7}
T5-11B-\textit{w/o Explanation} & 20\% & 4.5\% & 20\% & 0\% & 14\% & 0\% \\
DeBERTa-\textit{w/o Explanation} & 18\% & 0\% & 18\% & 3.6\% & 16\% & 5.5\% \\
\midrule 
T5-11B-\textit{w/ Explanation} & 16\% & 8\% & 14\% & 0\% & 13\% & 0\% \\
DeBERTa-\textit{w/ Explanation} & 15\% & 2.3\% & 15\% & 3.6\% & 15\% & 0\% \\
\bottomrule
\end{tabular}%
}
\caption{Model error analysis over expert annotation data. $\rightarrow$ means the expert-annotated label is on the left side and the predicted label is on the right side, and $\leftarrow$ vice versa. Percentages in the table represent the error rate in each expert labeled category. The results demonstrate that the main confusion for the models exists in ``Not Using - Misuse'', ``Recovery - Addiction'', and ``Not Using - Addiction''. These problems surface in an asymmetrical pattern -- one mis-classification direction matters more in the confusion.} 
\label{tab:expert_err_analysis}
\end{table*}

\begin{table*}[tbp!]
\centering
\resizebox{0.8\textwidth}{!}{%
\begin{tabular}{@{}cccccc@{}}
\toprule
Dataset & Agreement & T5-11B (w/o expl) & T5-11B (w/ expl) & DeBERTa (w/o expl) & DeBERTa (w/ expl) \\ 
\midrule
\multirow{3}{*}{Novice} & Unanimous Consent        & 95.5 / 94.0   &   97.1 / 95.7  &  87.1 / 85.9 & 90.0 / 89.6    \\
& Arguable (Majority Vote)      &   68.0 / 66.7     &    72.5 / 71.3    & 62.5 / 60.8  &  73.8 / 75.2  \\ 
& Arguable (All Annotations)      & 84.0 / 81.4   & 92.5 / 91.8  & 86.3 / 84.3  & 91.3 / 92.1 \\ 
\midrule
\multirow{3}{*}{Expert (FirstBatch)} & Unanimous Consent        & 92.7 / 85.0  &   96.4 / 97.1  &  83.6 / 72.7  & 89.1 / 83.2   \\
& Arguable (Majority Vote)      &   60.3 / 59.9     &    64.0 / 66.0    & 56.9 / 58.2  &  65.0 / 66.2 \\ 
& Arguable (All Annotations)      & 84.6 / 83.3   & 86.8 / 85.8  & 82.5 / 82.7  & 86.9 / 87.3 \\ 
\bottomrule

\end{tabular}
}

\caption{Ablation study on dataset annotation disagreement. Results are presented in "Accuracy/F1" format.
``Unanimous Consent'': all annotators agree on the same label; ``Arguable (Majority Vote)'': annotators have some disagreements, and majority voting is used as the correct label; ``Arguable (All annotations)'': disagreements exist, and any annotator label is considered correct. We  observe that models perform better on data with unanimous agreement than on arguable data. }
\label{tab:ablation_uncertainty}
\end{table*}

\section{Error Analysis}
\label{sec: error_analysis}

To understand the challenges and limitations of our best models, we perform an error analysis.

\shortparagraph{Model Errors}
We compute the confusion matrices
for DeBERTa and T5-11B on the expert evaluation dataset, as it is arguably more reliable and contains more examples.  We generally find that both \textit{w/ Explanation} and \textit{w/o Explanation} models
struggle with confusion between \textbf{1) Not Using - Misuse; 2) Recovery - Addiction; and 3) Not Using - Addiction} (\cref{tab:expert_err_analysis}, some of which we also noticed in the disagreement among annotators.
We notice that these problems surface in a very asymmetrical  pattern -- one direction (e.g., ``Not Using'' $\rightarrow$ ``Misuse'') matters more in this confusion.
Recall that our focus is on self-disclosures, so if a post discusses misuse (either a question or someone else misusing behavior), the expert label is Not Using, which might be difficult for models to capture in some cases.
Adding explanations mostly helps the model by reducing the confusion on the `Not Using` and `Recovery` labels, the two dominant as well as the top-2 most difficult categories
in expert-annotated data.

\shortparagraph{Error Annotations}
To better understand why the model makes mistakes
we did a thorough fine-grained error case annotation for the T5-11B \textit{w/ Explanation} model.
When analyzing why our models misclassified the \textbf{Recovery} class, we notice that
``Recovery'' can be a long process, and it is very common for users
to express their eagerness to get opiates (47.4\%), and/or to talk about their history of addiction 
(21.1\%). 
There are also some hard cases, such as a post showing the patient undergoing repeated recovery-relapse-recovery cycles (the model predicts ``relapse'' in this case). 
When analyzing the cases where our model misclassified the \textbf{Not Using} label, several cases emerge: 1)
\textit{asking a question about use/misuse/addiction (57.69\%)} 
(e.g., ``how much [Drug 1] should I take to get high (safely)? Can I use it with [Drug 2]?'', or asking questions about whether using drugs for certain syndromes is legal in some states and how much they should use);
2) \textit{irrelevant topics (23.08\%)} ( ``Merry Christmas!''); 
3) \textit{Others' overdose (7.69\%)} (discussing the addiction of their friends or family members; 
since we are interested in self-disclosures, this is labeled as Not Using, but models fail to recognize such subtle differences);
and 4) \textit{other drugs/substances, not opiates (3.85\%)} (as we focus on OUD, these posts are labeled ``Not Using''). 

\shortparagraph{Influence of Dataset Annotation Uncertainty} As we have already seen in the previous sections, annotators found it difficult to annotate several edge cases, which in turn brings uncertainty in the final annotation. 
To investigate how such uncertainty influences model performance, we do a further ablation study to test the model performance on data with unanimous agreement (all annotators give the same label) (47\% on the novice test set, 44\% on the expert test set)\footnote{Since only the first batch of expert data contains more than two annotators, for this study we only report ablation for this expert-annotated dataset.} and data where some disagreement exists, although majority voting can be reached (we call it arguable data). 
For the latter, we consider as gold label either the majority vote or any label chosen by at least one annotator. The
results are shown in \cref{tab:ablation_uncertainty}.  

We notice
that: 1) models perform better on data with unanimous agreement than on arguable data
(15\%-32\%); 
2) given the difficulty of the annotation task, if we consider all annotators' labels as gold (Arguable, all annotations), we can see the model can improve (14\%-25\%); 3) by comparing the performance on the first batch of expert-annotated data and the 
novice-annotated data, our models achieve very similar performance on instances with unanimous agreement and also when considering all annotators' labels as gold.
In arguable cases with majority voting,  however, 
models trained on novice-annotated data cannot perform as well on expert test sets where experts cannot reach a unanimous agreement or where we do not consider all labels. 
This confirms the fact that
the disagreement among
annotators will
influence the model performance and roughly quantify the performance bottleneck 
resulted from using majority voting as the gold label.

\section{Related Research}
\shortparagraph{Machine Learning for Substance Use}
Machine learning methods’ application to substance use research is growing~\citep{bharat2021big}. Several studies have attempted to predict substance use treatment completion among individuals with substance use disorders~\citep{gottlieb2022machine, acion2017use, hasan2021machine}. This study takes advantage of anonymous data to identify treatment needs among individuals who may not currently be in formal substance use treatment. Researchers have also used natural language processing to identify substance misuse in electronic health records~\cite{afshar2019natural, riddick2022natural} and to classify substances involved in drug overdose deaths~\citep{goodman2022development}. \citet{maclean2015forum77} collect user-level data on a social platform, Forum 77, to build a CRF model predicting three phases of drug use: using, withdrawing, and recovering. Our work is different in several aspects: 1) we propose an annotation scheme grounded in research
on addiction and substance use that defines behaviors such as Medical Use, Misuse, Addiction, Recovery, Relapse (and Not Using), that enable us to code self-disclosures of such behaviors using both expert and novice annotators; 2) we develop explanation-infused accurate models to identify self-disclosure at the post level. These two innovations will enable future research on using these models for a reliable global, user-level analysis across time.

\shortparagraph{Learning from Explanations}
There have been works focusing on
learning from human-annotated explanations.
\citet{wiegreffe2021measuring} investigates how free-form explanations and predicted labels are associated and use it as a property to evaluate the faithfulness of explanations. Different from that, our work focuses more on the utility of extractive span-level explanations as an additional source of supervision in a high-stakes domain and further shows how the quality of explanations impacts inference time results \cite{sun2022investigating}. 
Similar to our work, 
\citet{carton-etal-2022-learn} leverages extractive explanations and shows a consistent trend 
that using explanations can improve model performance in reading comprehension. Our work is most similar to \citet{huang-etal-2021-exploring}, who noticed that the quality of explanations could have a huge impact on model performance and explore the utility of extractive explanations, and to \citet{sun2022investigating}, who perform similar studies using free-form explanations.

\shortparagraph{Understanding the OUD continuum}
Scientists have explained how opioids produce changes in brain structure and function that promote and sustain addiction and contribute to relapse~\citep{koob2010neurocircuitry, abuse2016neurobiology}. Now recognized as a chronic but treatable disease of the brain, OUD is characterized by clinically significant impairments in health and social function and influenced by genetic, developmental, behavioral, social, and environmental factors~\citep{volkow2016neurobiologic}.
The HEALing Communities Study implemented the Opioid-overdose Reduction Continuum of Care Approach (ORCCA) to reduce opioid-overdose deaths across the OUD continuum~\citep{winhusen2020opioid}.
Taking advantage of self-disclosures on community-based social media, as this study aims to do, could lead
to the development of
interventions that better address risks associated with OUD. 

\section{Conclusions}
We presented a novel task aimed to deepen our understanding of how people move across the OUD continuum: given a user's post in an opioid-related Reddit, predict whether it contains a self-disclosure of various phases of OUD.
We provided an annotation scheme grounded in research on addiction and substance use,
which enables us to code self-disclosures of such behaviors using both expert and novice annotators. Following the annotation scheme, we created a high-quality dataset annotated with class labels and text explanations. We presented several state-of-the-art explanation-infused models, showing they can achieve accurate results in identifying self-disclosures of use, misuse, addiction, recovery, and relapse. Accurate models will enable further research in this space by considering a global user-level analysis across time. Our error analysis showed that explanations could provide insights both into annotator disagreement and errors in model predictions. In addition, our findings shed light
on how annotation uncertainty impacts model performance.

\section*{Acknowledgements}
This research is supported by the National Institutes of Health (NIH), National Institute on Drug Abuse, grant  \#UM1DA049412.
We want to thank
our expert  and novice annotators. We also thank the anonymous reviewers and chairs for providing insightful and constructive feedback to make this work more solid.

\section*{Limitations}
This study’s results are not without limitations. The
anonymity of Reddit users does not allow us to characterize the demographics or geographic extent of the study population.
Moreover, the current study looks at identifying self-disclosures at the message level without taking a global (user-level) and temporal view. In our future work, we plan to apply our models to study users' posts in opioid-related Reddits and observe their behavior over time.
In addition, we will work on improving our models to both predict a label and provide a textual explanation for the prediction.

\section*{Ethical Considerations}
For our data collection and annotation, we have obtained IRB approval. The source data comes from Reddit (r/opiates, r/OpiatesRecovery and r/Drugs), and is thus publicly available and anonymous. In addition, we preprocess the data to additionally remove any potentially identifiable information (see Section \ref{collection}).   All data is kept secure and online
userIDs are not associated to the posts.
For the expert annotation we compensated the experts with \$20 per hour, and the novice annotators with \$15 per hour. 

Our intention of developing datasets and models for predicting the stages of opioid use disorder is to help health professionals and/or social workers to both understand personal experiences of people across the opioid used disorder continuum and potentially to identify people
that might be at risk of overdose. The inclusion of explanations both in the annotation and in the prediction of our models could help the health professional better assess the models predictions. We emphasize that our models should be used with a human in the loop — for example a medical professional, or a social worker, who can look at the predicted labels and the explanations to decide whether or not they seem
sensible. 
We note that because most of our data were collected from Reddit, a website with a known overall
demographic skew (towards young, white, American men
), our conclusions about what explanations are associated with various OUD stages
cannot necessarily be applied to
broader groups of people. This might be particularly acute for vulnerable populations such as people with opioid use disorder (OUD). We hope that
this research stimulates more work by the research
community to consider and model ways in which
different groups self-disclose their experiences with OUD.

\bibliography{naacl}

\begin{thebibliography}{47}
\expandafter\ifx\csname natexlab\endcsname\relax\def\natexlab#1{#1}\fi

\bibitem[{Abuse et~al.(2016)Abuse, US, of~the Surgeon General~(US
  et~al.}]{abuse2016neurobiology}
Substance Abuse, Mental Health Services~Administration US, Office of~the
  Surgeon General~(US, et~al. 2016.
\newblock The neurobiology of substance use, misuse, and addiction.
\newblock In \emph{Facing Addiction in America: The Surgeon General's Report on
  Alcohol, Drugs, and Health [Internet]}. US Department of Health and Human
  Services.

\bibitem[{Acion et~al.(2017)Acion, Kelmansky, van~der Laan, Sahker, Jones, and
  Arndt}]{acion2017use}
Laura Acion, Diana Kelmansky, Mark van~der Laan, Ethan Sahker, DeShauna Jones,
  and Stephan Arndt. 2017.
\newblock Use of a machine learning framework to predict substance use disorder
  treatment success.
\newblock \emph{PloS one}, 12(4):e0175383.

\bibitem[{Afshar et~al.(2019)Afshar, Phillips, Karnik, Mueller, To, Gonzalez,
  Price, Cooper, Joyce, and Dligach}]{afshar2019natural}
Majid Afshar, Andrew Phillips, Niranjan Karnik, Jeanne Mueller, Daniel To,
  Richard Gonzalez, Ron Price, Richard Cooper, Cara Joyce, and Dmitriy Dligach.
  2019.
\newblock Natural language processing and machine learning to identify alcohol
  misuse from the electronic health record in trauma patients: development and
  internal validation.
\newblock \emph{Journal of the American Medical Informatics Association},
  26(3):254--261.

\bibitem[{Anderson et~al.(2017)Anderson, Bell, Gilbert, Davidson, Winter,
  Barratt, Win, Painter, Menone, Sayegh et~al.}]{anderson2017using}
Laurie~S Anderson, Heidi~G Bell, Michael Gilbert, Julie~E Davidson, Christina
  Winter, Monica~J Barratt, Beta Win, Jeffery~L Painter, Christopher Menone,
  Jonathan Sayegh, et~al. 2017.
\newblock Using social listening data to monitor misuse and nonmedical use of
  bupropion: a content analysis.
\newblock \emph{JMIR public health and surveillance}, 3(1):e6174.

\bibitem[{Bharat et~al.(2021)Bharat, Hickman, Barbieri, and
  Degenhardt}]{bharat2021big}
Chrianna Bharat, Matthew Hickman, Sebastiano Barbieri, and Louisa Degenhardt.
  2021.
\newblock Big data and predictive modelling for the opioid crisis: existing
  research and future potential.
\newblock \emph{The Lancet Digital Health}, 3(6):e397--e407.

\bibitem[{Bouzoubaa et~al.(2023)Bouzoubaa, Young, and
  Rezapour}]{bouzoubaa2023exploring}
Layla Bouzoubaa, Jordyn Young, and Rezvaneh Rezapour. 2023.
\newblock Exploring the landscape of drug communities on reddit: A network
  study.
\newblock In \emph{Proceedings of the International Conference on Advances in
  Social Networks Analysis and Mining}, pages 558--565.

\bibitem[{Brown et~al.(2020)Brown, Mann, Ryder, Subbiah, Kaplan, Dhariwal,
  Neelakantan, Shyam, Sastry, Askell, Agarwal, Herbert{-}Voss, Krueger,
  Henighan, Child, Ramesh, Ziegler, Wu, Winter, Hesse, Chen, Sigler, Litwin,
  Gray, Chess, Clark, Berner, McCandlish, Radford, Sutskever, and
  Amodei}]{brown2020language}
Tom~B. Brown, Benjamin Mann, Nick Ryder, Melanie Subbiah, Jared Kaplan,
  Prafulla Dhariwal, Arvind Neelakantan, Pranav Shyam, Girish Sastry, Amanda
  Askell, Sandhini Agarwal, Ariel Herbert{-}Voss, Gretchen Krueger, Tom
  Henighan, Rewon Child, Aditya Ramesh, Daniel~M. Ziegler, Jeffrey Wu, Clemens
  Winter, Christopher Hesse, Mark Chen, Eric Sigler, Mateusz Litwin, Scott
  Gray, Benjamin Chess, Jack Clark, Christopher Berner, Sam McCandlish, Alec
  Radford, Ilya Sutskever, and Dario Amodei. 2020.
\newblock \href
  {https://proceedings.neurips.cc/paper/2020/hash/1457c0d6bfcb4967418bfb8ac142f64a-Abstract.html}
  {Language models are few-shot learners}.
\newblock In \emph{Advances in Neural Information Processing Systems 33: Annual
  Conference on Neural Information Processing Systems 2020, NeurIPS 2020,
  December 6-12, 2020, virtual}.

\bibitem[{Bunting et~al.(2021)Bunting, Frank, Arshonsky, Bragg, Friedman, and
  Krawczyk}]{Bunting2021SociallysupportiveNA}
Amanda~M. Bunting, David Frank, Joshua Arshonsky, Marie~A. Bragg, Samuel~R.
  Friedman, and Noa Krawczyk. 2021.
\newblock Socially-supportive norms and mutual aid of people who use opioids:
  An analysis of reddit during the initial covid-19 pandemic.
\newblock \emph{Drug and alcohol dependence}, page 108672.

\bibitem[{Carton et~al.(2022)Carton, Kanoria, and Tan}]{carton-etal-2022-learn}
Samuel Carton, Surya Kanoria, and Chenhao Tan. 2022.
\newblock \href {https://doi.org/10.18653/v1/2022.findings-acl.86} {What to
  learn, and how: {T}oward effective learning from rationales}.
\newblock In \emph{Findings of the Association for Computational Linguistics:
  ACL 2022}, pages 1075--1088, Dublin, Ireland. Association for Computational
  Linguistics.

\bibitem[{Chan et~al.(2015)Chan, Lopez, and Sarkar}]{chan2015canary}
Brian Chan, Andrea Lopez, and Urmimala Sarkar. 2015.
\newblock The canary in the coal mine tweets: social media reveals public
  perceptions of non-medical use of opioids.
\newblock \emph{PloS one}, 10(8):e0135072.

\bibitem[{Choudhury and De(2014)}]{Choudhury2014MentalHD}
Munmun~De Choudhury and Sushovan De. 2014.
\newblock Mental health discourse on reddit: Self-disclosure, social support,
  and anonymity.
\newblock In \emph{ICWSM}.

\bibitem[{Cohen(1960)}]{cohen1960coefficient}
Jacob Cohen. 1960.
\newblock A coefficient of agreement for nominal scales.
\newblock \emph{Educational and psychological measurement}, 20(1):37--46.

\bibitem[{Degenhardt et~al.(2019)Degenhardt, Grebely, Stone, Hickman,
  Vickerman, Marshall, Bruneau, Altice, Henderson, Rahimi-Movaghar
  et~al.}]{degenhardt2019global}
Louisa Degenhardt, Jason Grebely, Jack Stone, Matthew Hickman, Peter Vickerman,
  Brandon~DL Marshall, Julie Bruneau, Frederick~L Altice, Graeme Henderson,
  Afarin Rahimi-Movaghar, et~al. 2019.
\newblock Global patterns of opioid use and dependence: harms to populations,
  interventions, and future action.
\newblock \emph{The Lancet}, 394(10208):1560--1579.

\bibitem[{El-Bassel et~al.(2022)El-Bassel, Hochstatter, Slavin, Yang, Zhang,
  and Muresan}]{el2022harnessing}
Nabila El-Bassel, Karli~R Hochstatter, Melissa~N Slavin, Chenghao Yang, Yudong
  Zhang, and Smaranda Muresan. 2022.
\newblock Harnessing the power of social media to understand the impact of
  covid-19 on people who use drugs during lockdown and social distancing.
\newblock \emph{Journal of addiction medicine}, 16(2):e123.

\bibitem[{Fleiss(1971)}]{fleiss1971measuring}
Joseph~L Fleiss. 1971.
\newblock Measuring nominal scale agreement among many raters.
\newblock \emph{Psychological bulletin}, 76(5):378.

\bibitem[{Goodman-Meza et~al.(2022)Goodman-Meza, Shover, Medina, Tang, Shoptaw,
  and Bui}]{goodman2022development}
David Goodman-Meza, Chelsea~L Shover, Jesus~A Medina, Amber~B Tang, Steven
  Shoptaw, and Alex~AT Bui. 2022.
\newblock Development and validation of machine models using natural language
  processing to classify substances involved in overdose deaths.
\newblock \emph{JAMA network open}, 5(8):e2225593--e2225593.

\bibitem[{Gottlieb et~al.(2022)Gottlieb, Yatsco, Bakos-Block, Langabeer, and
  Champagne-Langabeer}]{gottlieb2022machine}
Assaf Gottlieb, Andrea Yatsco, Christine Bakos-Block, James~R Langabeer, and
  Tiffany Champagne-Langabeer. 2022.
\newblock Machine learning for predicting risk of early dropout in a recovery
  program for opioid use disorder.
\newblock In \emph{Healthcare}, volume~10, page 223. MDPI.

\bibitem[{Hanson et~al.(2013{\natexlab{a}})Hanson, Burton, Giraud-Carrier,
  West, Barnes, and Hansen}]{hanson2013tweaking}
Carl~L Hanson, Scott~H Burton, Christophe Giraud-Carrier, Josh~H West,
  Michael~D Barnes, and Bret Hansen. 2013{\natexlab{a}}.
\newblock Tweaking and tweeting: exploring twitter for nonmedical use of a
  psychostimulant drug (adderall) among college students.
\newblock \emph{Journal of medical Internet research}, 15(4):e2503.

\bibitem[{Hanson et~al.(2013{\natexlab{b}})Hanson, Cannon, Burton, and
  Giraud-Carrier}]{hanson2013exploration}
Carl~Lee Hanson, Ben Cannon, Scott Burton, and Christophe Giraud-Carrier.
  2013{\natexlab{b}}.
\newblock An exploration of social circles and prescription drug abuse through
  twitter.
\newblock \emph{Journal of medical Internet research}, 15(9):e2741.

\bibitem[{Hasan et~al.(2021)Hasan, Young, Shi, Mohite, Young, Weiner
  et~al.}]{hasan2021machine}
Md~Mahmudul Hasan, Gary~J Young, Jiesheng Shi, Prathamesh Mohite, Leonard~D
  Young, Scott~G Weiner, et~al. 2021.
\newblock A machine learning based two-stage clinical decision support system
  for predicting patients’ discontinuation from opioid use disorder
  treatment: Retrospective observational study.
\newblock \emph{BMC Medical Informatics and Decision Making}, 21(1):1--21.

\bibitem[{He et~al.(2021)He, Gao, and Chen}]{he2021debertav3}
Pengcheng He, Jianfeng Gao, and Weizhu Chen. 2021.
\newblock \href {https://arxiv.org/abs/2111.09543} {Debertav3: Improving
  deberta using electra-style pre-training with gradient-disentangled embedding
  sharing}.
\newblock \emph{arXiv preprint arXiv:2111.09543}.

\bibitem[{Hu et~al.(2019)Hu, Phan, Geller, Iezzi, Vo, Dou, and
  Chun}]{hu2019ensemble}
Han Hu, NhatHai Phan, James Geller, Stephen Iezzi, Huy~T Vo, Dejing Dou, and
  Soon~Ae Chun. 2019.
\newblock An ensemble deep learning model for drug abuse detection in sparse
  twitter-sphere.
\newblock In \emph{MedInfo}, pages 163--167.

\bibitem[{Huang et~al.(2021)Huang, Zhu, Feng, and
  Zhao}]{huang-etal-2021-exploring}
Quzhe Huang, Shengqi Zhu, Yansong Feng, and Dongyan Zhao. 2021.
\newblock \href {https://doi.org/10.18653/v1/2021.acl-long.433} {Exploring
  distantly-labeled rationales in neural network models}.
\newblock In \emph{Proceedings of the 59th Annual Meeting of the Association
  for Computational Linguistics and the 11th International Joint Conference on
  Natural Language Processing (Volume 1: Long Papers)}, pages 5571--5582,
  Online. Association for Computational Linguistics.

\bibitem[{Koob and Volkow(2010)}]{koob2010neurocircuitry}
George~F Koob and Nora~D Volkow. 2010.
\newblock Neurocircuitry of addiction.
\newblock \emph{Neuropsychopharmacology}, 35(1):217--238.

\bibitem[{Krausz et~al.(2021)Krausz, Westenberg, and Ziafat}]{krausz2021opioid}
R~Michael Krausz, Jean~Nicolas Westenberg, and Kimia Ziafat. 2021.
\newblock The opioid overdose crisis as a global health challenge.
\newblock \emph{Current Opinion in Psychiatry}, 34(4):405--412.

\bibitem[{Lampinen et~al.(2022)Lampinen, Dasgupta, Chan, Matthewson, Tessler,
  Creswell, McClelland, Wang, and Hill}]{lampinen2022can}
Andrew~K Lampinen, Ishita Dasgupta, Stephanie~CY Chan, Kory Matthewson,
  Michael~Henry Tessler, Antonia Creswell, James~L McClelland, Jane~X Wang, and
  Felix Hill. 2022.
\newblock \href {https://arxiv.org/abs/2204.02329} {Can language models learn
  from explanations in context?}
\newblock \emph{arXiv preprint arXiv:2204.02329}.

\bibitem[{Loshchilov and Hutter(2018)}]{loshchilov2018fixing}
Ilya Loshchilov and Frank Hutter. 2018.
\newblock Fixing weight decay regularization in adam.

\bibitem[{MacLean et~al.(2015)MacLean, Gupta, Lembke, Manning, and
  Heer}]{maclean2015forum77}
Diana MacLean, Sonal Gupta, Anna Lembke, Christopher Manning, and Jeffrey Heer.
  2015.
\newblock Forum77: An analysis of an online health forum dedicated to addiction
  recovery.
\newblock In \emph{Proceedings of the 18th ACM Conference on Computer Supported
  Cooperative Work \& Social Computing}, pages 1511--1526.

\bibitem[{NIDA(2007)}]{nida2007drugs}
NIDA. 2007.
\newblock \emph{Drugs, brains, and behavior: The science of addiction}.
\newblock National Institute on Drug Abuse, National Institutes of Health, US.

\bibitem[{OpenAI(2023)}]{OpenAI2023GPT4TR}
OpenAI. 2023.
\newblock Gpt-4 technical report.
\newblock \emph{ArXiv}, abs/2303.08774.

\bibitem[{Pandrekar et~al.(2018)Pandrekar, Chen, Gopalkrishna, Srivastava,
  Saltz, Saltz, and Wang}]{pandrekar2018social}
Sheetal Pandrekar, Xin Chen, Gaurav Gopalkrishna, Avi Srivastava, Mary Saltz,
  Joel Saltz, and Fusheng Wang. 2018.
\newblock Social media based analysis of opioid epidemic using reddit.
\newblock In \emph{AMIA Annual Symposium Proceedings}, volume 2018, page 867.
  American Medical Informatics Association.

\bibitem[{Park and Conway(2017)}]{park2017tracking}
Albert Park and Mike Conway. 2017.
\newblock Tracking health related discussions on reddit for public health
  applications.
\newblock In \emph{AMIA annual symposium proceedings}, volume 2017, page 1362.
  American Medical Informatics Association.

\bibitem[{Park et~al.(2020)Park, Rouhani, Beletsky, Vincent, Saloner, and
  Sherman}]{park2020situating}
Ju~Nyeong Park, Saba Rouhani, Leo Beletsky, Louise Vincent, Brendan Saloner,
  and Susan~G Sherman. 2020.
\newblock Situating the continuum of overdose risk in the social determinants
  of health: a new conceptual framework.
\newblock \emph{The Milbank Quarterly}, 98(3):700--746.

\bibitem[{Phan et~al.(2017)Phan, Chun, Bhole, and Geller}]{phan2017enabling}
Nhathai Phan, Soon~Ae Chun, Manasi Bhole, and James Geller. 2017.
\newblock Enabling real-time drug abuse detection in tweets.
\newblock In \emph{2017 IEEE 33rd international conference on data engineering
  (ICDE)}, pages 1510--1514. IEEE.

\bibitem[{Raffel et~al.(2020)Raffel, Shazeer, Roberts, Lee, Narang, Matena,
  Zhou, Li, and Liu}]{raffel2020exploring}
Colin Raffel, Noam Shazeer, Adam Roberts, Katherine Lee, Sharan Narang, Michael
  Matena, Yanqi Zhou, Wei Li, and Peter~J Liu. 2020.
\newblock Exploring the limits of transfer learning with a unified text-to-text
  transformer.
\newblock \emph{Journal of Machine Learning Research}, 21:1--67.

\bibitem[{Rasley et~al.(2020)Rasley, Rajbhandari, Ruwase, and
  He}]{rasley2020deepspeed}
Jeff Rasley, Samyam Rajbhandari, Olatunji Ruwase, and Yuxiong He. 2020.
\newblock \href {https://dl.acm.org/doi/10.1145/3394486.3406703} {Deepspeed:
  System optimizations enable training deep learning models with over 100
  billion parameters}.
\newblock In \emph{{KDD} '20: The 26th {ACM} {SIGKDD} Conference on Knowledge
  Discovery and Data Mining, Virtual Event, CA, USA, August 23-27, 2020}, pages
  3505--3506. {ACM}.

\bibitem[{Riddick and Choo(2022)}]{riddick2022natural}
Tyne~A Riddick and Esther~K Choo. 2022.
\newblock Natural language processing to identify substance misuse in the
  electronic health record.
\newblock \emph{The Lancet Digital Health}, 4(6):e401--e402.

\bibitem[{Sanh et~al.(2022)Sanh, Webson, Raffel, Bach, Sutawika, Alyafeai,
  Chaffin, Stiegler, Le~Scao, Raja et~al.}]{sanh2022multitask}
Victor Sanh, Albert Webson, Colin Raffel, Stephen Bach, Lintang Sutawika, Zaid
  Alyafeai, Antoine Chaffin, Arnaud Stiegler, Teven Le~Scao, Arun Raja, et~al.
  2022.
\newblock Multitask prompted training enables zero-shot task generalization.
\newblock In \emph{The Tenth International Conference on Learning
  Representations}.

\bibitem[{Smith et~al.(2013)Smith, Dart, Katz, Paillard, Adams, Comer, Degroot,
  Edwards, Haddox, Jaffe et~al.}]{smith2013classification}
Shannon~M Smith, Richard~C Dart, Nathaniel~P Katz, Florence Paillard, Edgar~H
  Adams, Sandra~D Comer, Aldemar Degroot, Robert~R Edwards, J~David Haddox,
  Jerome~H Jaffe, et~al. 2013.
\newblock Classification and definition of misuse, abuse, and related events in
  clinical trials: Acttion systematic review and recommendations.
\newblock \emph{Pain{\textregistered}}, 154(11):2287--2296.

\bibitem[{Sumner et~al.(2019)Sumner, Galik, Mathieu, Ward, Kiley, Bartholow,
  Dingwall, and Mork}]{Sumner2019TemporalAG}
Steven~A. Sumner, Stacey Galik, Jennifer Mathieu, Megan Ward, Thomas~R. Kiley,
  Bradford~N. Bartholow, Alison Dingwall, and Peter Mork. 2019.
\newblock Temporal and geographic patterns of social media posts about an
  emerging suicide game.
\newblock \emph{The Journal of adolescent health : official publication of the
  Society for Adolescent Medicine}, 65 1:94--100.

\bibitem[{Sun et~al.(2022)Sun, Swayamdipta, May, and Ma}]{sun2022investigating}
Jiao Sun, Swabha Swayamdipta, Jonathan May, and Xuezhe Ma. 2022.
\newblock \href {https://arxiv.org/abs/2206.11083} {Investigating the benefits
  of free-form rationales}.
\newblock \emph{arXiv preprint arXiv:2206.11083}.

\bibitem[{USDHHS(2017)}]{us2017hhs}
USDHHS. 2017.
\newblock Hhs acting secretary declares public health emergency to address
  national opioid crisis.
\newblock \emph{Washington, DC: USDHHS}.

\bibitem[{Volkow(2007)}]{volkow2007science}
ND~Volkow. 2007.
\newblock How science has revolutionized the understanding of drug addiction.
\newblock \emph{Drugs, Brains and Behavior: The Science of Addiction}.

\bibitem[{Volkow et~al.(2016)Volkow, Koob, and
  McLellan}]{volkow2016neurobiologic}
Nora~D Volkow, George~F Koob, and A~Thomas McLellan. 2016.
\newblock Neurobiologic advances from the brain disease model of addiction.
\newblock \emph{New England Journal of Medicine}, 374(4):363--371.

\bibitem[{Wiegreffe et~al.(2021)Wiegreffe, Marasovi{\'c}, and
  Smith}]{wiegreffe2021measuring}
Sarah Wiegreffe, Ana Marasovi{\'c}, and Noah~A Smith. 2021.
\newblock Measuring association between labels and free-text rationales.
\newblock In \emph{Proceedings of the 2021 Conference on Empirical Methods in
  Natural Language Processing}, pages 10266--10284.

\bibitem[{Winhusen et~al.(2020)Winhusen, Walley, Fanucchi, Hunt, Lyons,
  Lofwall, Brown, Freeman, Nunes, Beers et~al.}]{winhusen2020opioid}
Theresa Winhusen, Alexander Walley, Laura~C Fanucchi, Tim Hunt, Mike Lyons,
  Michelle Lofwall, Jennifer~L Brown, Patricia~R Freeman, Edward Nunes, Donna
  Beers, et~al. 2020.
\newblock The opioid-overdose reduction continuum of care approach (orcca):
  evidence-based practices in the healing communities study.
\newblock \emph{Drug and alcohol dependence}, 217:108325.

\bibitem[{Wolf et~al.(2020)Wolf, Debut, Sanh, Chaumond, Delangue, Moi, Cistac,
  Rault, Louf, Funtowicz, Davison, Shleifer, von Platen, Ma, Jernite, Plu, Xu,
  Le~Scao, Gugger, Drame, Lhoest, and Rush}]{wolf-etal-2020-transformers}
Thomas Wolf, Lysandre Debut, Victor Sanh, Julien Chaumond, Clement Delangue,
  Anthony Moi, Pierric Cistac, Tim Rault, Remi Louf, Morgan Funtowicz, Joe
  Davison, Sam Shleifer, Patrick von Platen, Clara Ma, Yacine Jernite, Julien
  Plu, Canwen Xu, Teven Le~Scao, Sylvain Gugger, Mariama Drame, Quentin Lhoest,
  and Alexander Rush. 2020.
\newblock \href {https://doi.org/10.18653/v1/2020.emnlp-demos.6} {Transformers:
  State-of-the-art natural language processing}.
\newblock In \emph{Proceedings of the 2020 Conference on Empirical Methods in
  Natural Language Processing: System Demonstrations}, pages 38--45, Online.
  Association for Computational Linguistics.

\end{thebibliography}

\appendix

\begin{table*}[!ht]
\centering
\resizebox{0.95\textwidth}{!}{%
\renewcommand{\arraystretch}{1.05}
\begin{tabular}{|c|l|}
\hline
Medical Use
& \begin{tabular}[c]{@{}l@{}}Medical use is defined as the use of prescription opioids that were prescribed by a medical\\ professional for the purpose of treating a medical condition\end{tabular}                                                              \\\hline
Misuse                                                 & \begin{tabular}[c]{@{}l@{}}Misuse is defined as the use of a substance that does not follow medical indications or prescribed \\ dosing. Substances are commonly used for nontherapeutic purposes to obtain psychotropic (eg, \\ euphoric, seditative, or anxiolytic) effects. Misuse is not restricted to prescription opioids.\end{tabular}                                \\ \hline
Addiction                                              & \begin{tabular}[c]{@{}l@{}}Addiction is defined as compulsive opioid use that occurs despite personal harm or negative \\ consequences. Addiction may involve impaired control and craving, neurobiologic dysfunction,\\ physical and psychological dependence, and withdrawal.\end{tabular}                                                                 \\ \hline
Recovery                                               & \begin{tabular}[c]{@{}l@{}}Recovery is a process of change through which individuals improve their health and wellness,\\ live a self-directed life, and strive to reach their full potential without using opioids.\end{tabular}                                                                 \\ \hline
Relapse                                                & Relapse is defined as the return to opioid use after an attempt to quit.
         \\ \hline
Not Using
& \begin{tabular}[c]{@{}l@{}}Posts should be labeled 'Not Using' which are about substances other than opioids \\ (e.g.,
marijuana), another person who uses opioids (e.g., family or friend), general questions \\ about opioids without evidence that the persons use opioids, and irrelevant information.
\end{tabular} \\ \hline
\end{tabular}
}
\caption{Expert guidelines on how to assign each post one of the six stages of the OUD continuum} 
\label{guidelines}
\end{table*}

\section{Annotation Guideline}
\label{app: annotation_guideline}
A brief annotation guideline created by experts is shown in \cref{guidelines}, which explains the definition for each OUD category. This guideline also comes with example posts picked by experts that help annotator under the definitions and we show them in \cref{tab:guideline_examples}. The full guideline is too large to put in this paper so we will release it in our GitHub project. 

Experts also help draft FAQs for clarification in the initial trial of annotations. Examples of FAQs are shown below:
\begin{dialogue}
    \speak{Question} What if the post described family, friend, or peer opioid use and there is no evidence that the person posting used opioids?
    \speak{Answer} This post should be labeled ‘not using’ because there is no evidence that the individual posting the comment used opioids.
    \speak{Question} What if the post discusses using stimulants, marijuana, or other drugs that are not opioids?
    \speak{Answer} This post should be labeled ‘not using’ because this study is specifically focused on understanding the development and advancement of opioid use disorder.
    \speak{Question} Is ‘misuse’ restricted to prescription opioids?
    \speak{Answer} We have decided for the purpose of this study that misuse will NOT be restricted to prescription opioids. Therefore, if someone describes trying a synthetic or semi-synthetic opioid (e.g., heroin) or using it infrequently, but does not display signs of being addicted, this post should be labeled ‘misuse.’
    \speak{Question} What if the post asks a question about opioid use, but does not provide evidence that the individual posting the comment used opioids?
    \speak{Answer} This post should be labelled ‘not using’ because there is no evidence that the individual posting the comment used opioids. They may just be curious.
    \speak{Question} If someone reports using drugs that are NOT opioids during a period of time when they are attempting to quit (i.e., when they are in recovery), should this be considered ‘relapse?’
    \speak{Answer} Because this study is focused on opioid use disorder, we have defined relapse as use of opioids after an attempt to quit. Thus, if the individual used other drugs that are not opioids during recovery, we will not consider this relapse.
\end{dialogue}

\begin{table*}[!ht]
\centering
\resizebox{0.95\textwidth}{!}{%
\renewcommand{\arraystretch}{1.05}
\begin{tabular}{|c|l|}
\hline
Medical Use
& \begin{tabular}[c]{@{}l@{}}I got pretty decent surgery on my feet and was prescribed 400 mg of oxy after takeing that\\ In about 10 days as needed due to pain ( never takeing more then prescribed ) \\but I have had minor withdrawal symptoms I took a 3 day break \\when do you think i can start taking it agian when my foot hurts and not withdrawal\end{tabular}                                                              \\\hline
Misuse                                                 & \begin{tabular}[c]{@{}l@{}}So I was given vicoprofen (7.5 hydrocodone to 200tylenol) for a severe toothache. \\ I have been using it as prescribed but dumb ass me decided to take quite a large dose last night after missing a few normal doses. \\ If I go back to using the normal doses now, after one large one, is it still going to be effective? \\Or should I wait and if so how long."\end{tabular}                                \\ \hline
Addiction                                              & \begin{tabular}[c]{@{}l@{}}I have been on opiates (oxycodone/contin) for like 5-6 years.\\ Started off really small, got really big, now at like medium use- compared to before. \\ I spent the last year or so very slowly tapering from my high of 330mg/day to now about 80mg/day. \\At this point is just maintenance to be able to function properly in my everyday life w out being sick or too tired.\end{tabular}                                                                 \\ \hline
Recovery                                               & \begin{tabular}[c]{@{}l@{}}"7 days clean from heroin today after having been IV'ing it on my daily basis since August, 2020"\end{tabular}                                                                 \\ \hline
Relapse                                                &\begin{tabular}[c]{@{}l@{}} "i made it 70 days clean. now i'm back to square one. \\ i wish i could stop but i can't. now i'm shooting 2 grams a day, plus 2-4 grams of coke a day. \\everytime i relapse i get more and more addicted. anyone else experience this ? that when you relapse it gets more out of control.\\ but godam i love it, i love the feeling, the lifestyle. "\end{tabular}
         \\ \hline
Not Using
& \begin{tabular}[c]{@{}l@{}}"How do you feel about Oxford houses/halfway houses/sober houses?"\\ "Dreary , rainy day here , thought about using , now binge watching Reno 911 instead . It’s so funny lol
\end{tabular} \\ \hline
\end{tabular}
}
\caption{Expert guidelines on example posts for each category} 
\label{tab:guideline_examples}
\end{table*}

\begin{table*}[htbp!]
\centering
\resizebox{0.90\textwidth}{!}{
\begin{tabular}{lcccc}
\toprule
\textbf{Category} & \textbf{T5-11B w/ Explanation} & \textbf{T5-11B w/o Explanation} & \textbf{GPT-4 w/ Explanation} & \textbf{GPT-4 w/o Explanation} \\
\midrule
Addiction & \(88\% / 98\%\) & \(84\% / 94\%\) & \(28\% / 62\%\) & \(20\% / 49\%\) \\
Medical Use & \(84\% / 87\%\) & \(76\% / 87\%\) & \(88\% / 87\%\) & \(84\% / 87\%\) \\
Misuse & \(88\% / 70\%\) & \(88\% / 75\%\) & \(76\% / 89\%\) & \(72\% / 82\%\) \\
Not Using & \(68\% / 66\%\) & \(68\% / 63\%\) & \(36\% / 30\%\) & \(32\% / 27\%\) \\
Recovery & \(92\% / 83\%\) & \(84\% / 67\%\) & \(88\% / 81\%\) & \(84\% / 70\%\) \\
Relapse & \(84\% / 81\%\) & \(88\% / 69\%\) & \(88\% / 81\%\) & \(88\% / 69\%\) \\
\bottomrule
\end{tabular}
}
\caption{Class-wise performance decomposition for different models. Results are presented in a format of ``Accuracy on Novice Test Set/Accuracy on Expert Test Set''.} 
\label{tab:class-wise-perf}
\end{table*}

\section{Heatmap for Worker-Expert labels over the Qualification Test}
\label{app:heatmap}
\begin{figure}
    \centering
        \centering
        \includegraphics[trim={1cm 2cm 1cm 2cm},clip,width=\columnwidth]{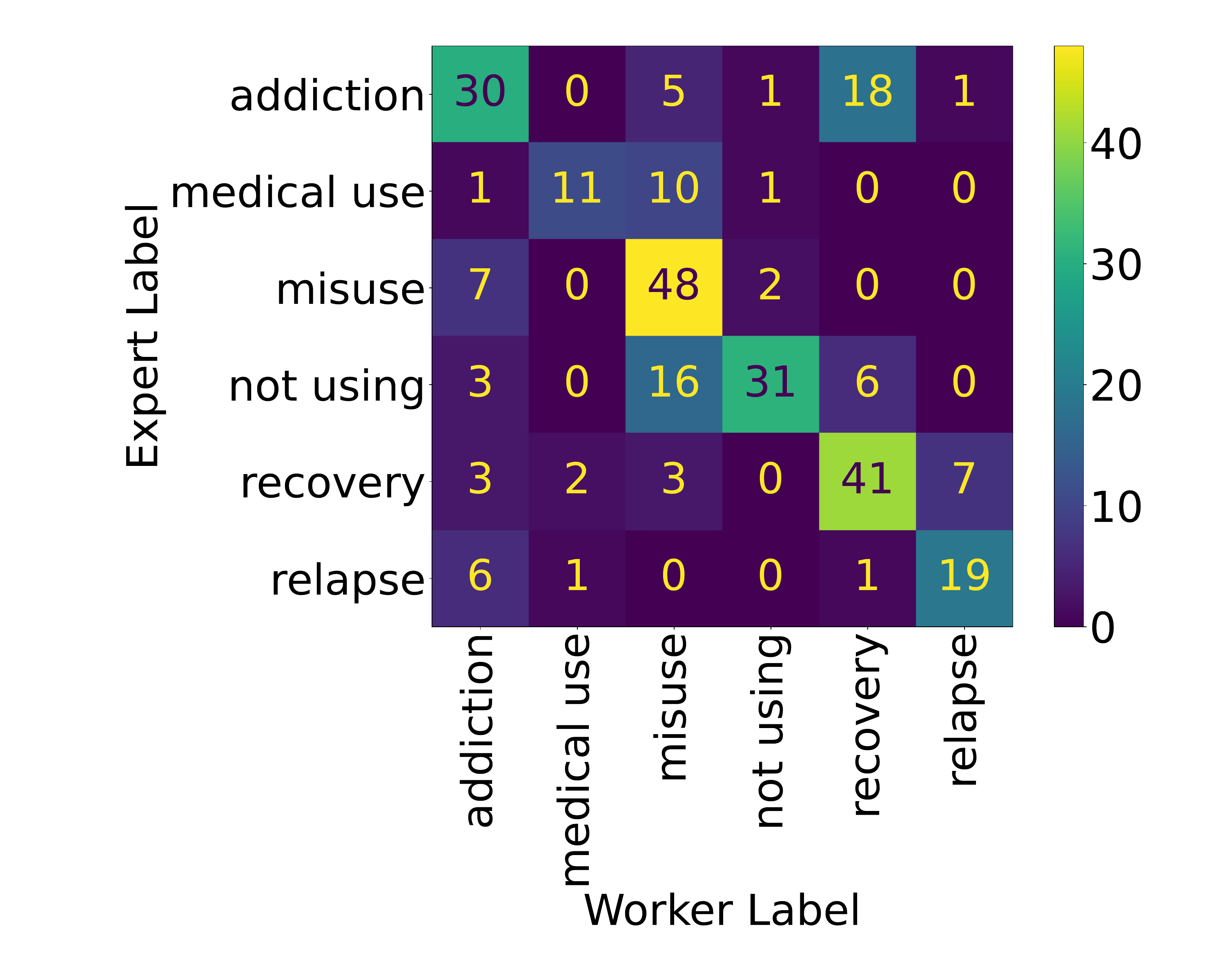}
        \caption{Heatmap for all worker labels and expert labels over the qualification test. }
        \label{fig: worker-expert-disagree}
        \vspace{-5mm}
\end{figure}

The heatmap summarizes the difference in annotations between workers and experts over the qualification test is shown in \cref{fig: worker-expert-disagree}.

\section{In-Context Samples in Few-Shot Learning Settings}
\label{app:in-context-samples}
The 13 in-context samples we used for prompting in the few-shot learning setting are shown in \cref{tab:in-context-samples}. These in-context samples are selected as the representative samples under each category after discussions with experts providing the annotations. The distribution of classes is decided based on preliminary experiments on held-out data. 

    \begin{table*}[h!]
    \centering
    \resizebox{0.98\textwidth}{!}{
    \begin{tabular}{|l|l|l|l|}
    \hline
    \textbf{Title} & \textbf{Text} & \textbf{Opioid Usage Label} & \textbf{Explanation} \\
    \hline
    Advice welcome & I am a 23yr old female, been addicted to H for 3 y... & relapse & I am a 23yr old female, been addicted to H for 3 y... \\ \hline
Nearly threw two months down the drain today. & Well everyone I've been clean from Heroin for the ... & recovery & Well everyone I've been clean from Heroin for the ... \\ \hline
2weeks clean from all opiates.i just want to vent a bit. & So two weeks ago I quit my job, opiate use, and go... & recovery & 2weeks clean from all opiates \\ \hline
Listening to Christmas music.... & I could have been someone ''Well so could anyone..... & not using & I could have been someone ''Well so could anyone..... \\ \hline
Heroin use & Hi, non user here, just curious as to what heroin ... & not using & Hi, non user here, just curious as to what heroin ... \\ \hline
Anyone here either a lawyer or have solid solid knowledge or experience on drug & Not really comfortable discussing this publicly, b... & not using & Anyone here either a lawyer or have solid solid kn... \\ \hline
Supeudol oxycodone, sniffable? & My doctor changed my oxy prescription to supeudol ... & medical use & My doctor changed my oxy prescription to supeudol ... \\ \hline
Back in the cycle.. & I started using more again (daily when I can) afte... & addiction & I started using more again (daily when I can) afte... \\ \hline
ROA 30mg roxis (blues) & I am currently on 7 blues. That i have done over t... & misuse & I am currently on 7 blues. That i have done over t... \\ \hline
Hey guys! Kinda worried! & Hey, I took abour 4 lines of heroin at 6pmAnother ... & misuse & Hey, I took abour 4 lines of heroin at 6pmAnother ... \\ \hline
Really wanting to try heroin :/ & I just wanna say before I start, Ik how bad it is ... & misuse & I’ve used weed, Xanax, coke, I’m off of 2 Kpins ri... \\ \hline

    \end{tabular}
    }

\caption{Thirteen in-context examples for each
Opioid Usage category. }
    \label{tab:in-context-samples}
    \end{table*}

\section{Post-processing needed for processing GPT-3 outputs}
\label{app:post_processing_gpt3}
In our experiments, we generally find that GPT-3 outputs cannot be taken as exact match as outputs and can contain some typos, we provide the following post-processing for it:

\begin{enumerate}
    \item We ignore any content after a newline symbol (i.e., ``$\backslash n$ '').
    \item If GPT3 responses are like ``1) ... 2) ...'', we only take the term between ``1)'' and ``2)''. 
    \item For morphological changes like predicting ``misuse'' as ``misusing'', we manually recover these changes. 
    \item For typos like ``misue'', we would manually correct it to be ``misuse''. 
\end{enumerate}

We tried to apply the same processing for GPT-4 as well, but we did not find significant changes. This may indicate GPT-4 has better instruction-following capability while GPT-3 does not.

\section{Fully Supervised Fine-Tuning Details}
\label{sec:fully_supervised_fine-tuning_detail}
In this section, we give details for fine-tuning language models under fully supervised setting.

For fine-tuning DeBERTa-v3-large~\citep{he2021debertav3}, we adopt the widely-used huggingface transformers fine-tuning implementation~\citep{wolf-etal-2020-transformers} with the learning rate of $2e-5$ and fine-tune the model for $10$ epochs. For optimizer, we use AdamW~\citep{loshchilov2018fixing}. 

For fine-tuning T5~\cite{raffel2020exploring}, we adopt the huggingface transformers implementation ~\citep{wolf-etal-2020-transformers} to fine-tune two versions of T5, the 3B model and the 11B model, respectively. We hold out 100 examples for validation from our training set to tune our models and find the best checkpoint. We use a batch size of 1024 for the 3B model and 512 for the 11B model. Further, we maintain a learning rate of 1e-4 and AdamW optimizer~\citep{loshchilov2018fixing} for both 3B and 11B models. We fine-tune all models on 4 A100 GPUs and use Deepspeed \cite{rasley2020deepspeed} integration for the 11B model. We fine-tune the 3B model for 20 epochs and the 11B model for eight epochs. During fine-tuning, we restrict the maximum sequence length of the source to 1024 (via truncation), while our target length is less than the default 128 tokens.

\section{Class-Wise Performance Decomposition}
\label{app:class-wise-main-exp}
In \cref{sec: exp}, we show the model average performance w/ and w/o explanations over all categories in \cref{tab:main_exp}. As there exist significant differences between OUD categories and their individual importance can vary depending on application purposes, we further show the class-wise performance decomposition in \cref{tab:class-wise-perf} for both expert and novice annotated test sets. 

\section{Scientific Artifacts}
In this paper, we use the following artifacts: 

\textit{cleantext}\footnote{\url{www.github.com/prasanthg3/cleantext}} (v1.1.4): is an open-source python package to clean raw text data. We use it to preprocess raw social media posts. This toolkit is released under an MIT license. 

\textit{Transformers} \cite{wolf-etal-2020-transformers}\footnote{https://github.com/huggingface/transformers} (v4.35.0): provides thousands of pretrained models to perform tasks on different modalities such as text, vision, and audio. We use it for model training and inference. This toolkit is released under an Apache-2.0 license. 

\textit{OpenAI-python}\footnote{\url{https://github.com/openai/openai-python}} (v1.0.0): provides convenient access to the OpenAI REST API from any Python 3.7+ application. The library includes type definitions for all request params and response fields, and offers both synchronous and asynchronous clients powered by httpx. We use it for prompting the GPT-series models. This toolkit is released under an Apache-2.0 license.

In addition, we plan to release our codebase and dataset under an MIT license in the formal version. 

\end{document}